\documentclass[8pt, twocolumn, letterpaper]{article}

\usepackage[utf8]{inputenc}
\usepackage{times}
\usepackage{epsfig}
\usepackage{graphicx}
\usepackage{amsmath}
\usepackage{amssymb}
\usepackage{booktabs}
\usepackage{multirow}
\usepackage{algorithm}
\usepackage{algorithmic}
\usepackage{url}
\usepackage[breaklinks=true,bookmarks=false]{hyperref}

\title{\textbf{DiMEx: Breaking the Cold Start Barrier in Data-Free Model Extraction via Latent Diffusion Priors}}

\author{
  \textbf{Yash Thesia}\thanks{Corresponding author.} \\
  Dept. of Computer Science \\
  New York University \\
  New York, NY, USA \\ 
  \texttt{yt2188@nyu.edu}
\and
  \textbf{Meera Suthar} \\
  Dept. of Computer Science \\
  New York University \\
  New York, NY, USA \\ 
  \texttt{ms12418@nyu.edu}
}

\date{}

\begin{document}

\maketitle

\begin{abstract}
Model stealing attacks pose an existential threat to Machine Learning as a Service (MLaaS), allowing adversaries to replicate proprietary models for a fraction of their training cost. While Data-Free Model Extraction (DFME) has emerged as a stealthy vector, it remains fundamentally constrained by the ``Cold Start'' problem: Generative Adversarial Network (GAN)-based adversaries waste thousands of queries converging from random noise to meaningful data. We propose \textbf{DiMEx}, a framework that weaponizes the rich semantic priors of pre-trained Latent Diffusion Models to bypass this initialization barrier entirely. By employing Random Embedding Bayesian Optimization (REMBO) within the generator's latent space, DiMEx synthesizes high-fidelity queries immediately, achieving \textbf{52.1\%} agreement on Street View House Numbers (SVHN) with just 2,000 queries—outperforming state-of-the-art GAN baselines by over \textbf{16\%}. To counter this highly semantic threat, we introduce the \textbf{Hybrid Stateful Ensemble (HSE)} defense, which identifies the unique ``optimization trajectory'' of latent-space attacks. Our results demonstrate that while DiMEx evades static distribution detectors, HSE exploits this temporal signature to suppress attack success rates to \textbf{21.6\%} with negligible latency.
\end{abstract}
\paragraph{Keywords:} Data-free model extraction; model stealing attacks; latent diffusion models; cold start problem; query synthesis; Bayesian optimization; Hybrid Stateful Ensemble.

\section{Introduction}
\label{sec:intro}

Deep learning models have evolved into high-value intellectual property, forming the economic backbone of Machine Learning as a Service (MLaaS) platforms. However, this centralized deployment model exposes a critical vulnerability: \textit{Model Extraction} (or Model Stealing) attacks. By querying a victim model's black-box API and aggregating the predictions, an adversary can train a surrogate model that functionally replicates the proprietary victim~\cite{tramer2016stealing, papernot2017practical}, effectively bypassing the immense costs of data acquisition and training.

The efficacy of these attacks has traditionally been constrained by the availability of data. Early methods like \textit{Knockoff Nets}~\cite{orekondy2019knockoff} relied on ``proxy datasets''---public data semantically similar to the victim's training distribution. While effective, this assumption breaks down in specialized domains (e.g., medical or industrial imaging) where no such public proxy exists. Consequently, the field shifted toward \textit{Data-Free Model Extraction (DFME)}~\cite{truong2021data, kariyappa2021maze}, where adversaries train Generative Adversarial Networks (GANs) to synthesize queries from scratch.

However, we identify a fundamental bottleneck in current DFME methodologies: the \textbf{``Cold Start'' problem}~\cite{sanyal2022towards}. GAN-based generators are initialized with random weights, producing uninformative Gaussian noise in the early stages of the attack. Since the victim's decision boundaries are undefined for such Out-of-Distribution (OOD) noise, the returned gradients are sparse and noisy, forcing the adversary to waste thousands of valuable queries just to ``warm up'' the generator. This inefficiency is critical; in real-world scenarios where API queries are metered and monitored, the initial noise burst acts as a blatant signature for intrusion detection systems.

The urgency of this threat is amplified by the rise of Foundation Models (FMs). Recent analysis by Raj et al. (2025)~\cite{raj2025foundation} reveals that victims fine-tuned from FMs (e.g., Vision Transformers) are significantly easier to steal than traditional CNNs due to shared feature representations. Furthermore, adaptive attacks like \textit{Black-box Extraction of Side-channel Features (BESA)}~\cite{ren2025besa} have begun to bypass active defenses by recovering clean features from perturbed queries. This convergence of vulnerable victim models and sophisticated extraction techniques necessitates a paradigm shift in how we generate attack queries.

\textbf{Our Approach.} In this work, we propose \textbf{DiMEx} (Diffusion-based Model Extraction), a framework that eliminates the Cold Start problem by replacing the untrained GAN generator with the frozen, semantic prior of a \textit{Latent Diffusion Model (LDM)}. Instead of optimizing in pixel space, DiMEx employs Random Embedding Bayesian Optimization (REMBO) to search the compact latent manifold of Stable Diffusion. This ensures that \textit{every} query---even the very first---is a semantically valid image, providing rich, informative gradients immediately and bypassing the warm-up phase entirely.

Because DiMEx queries lie on the natural image manifold, they render traditional OOD-based defenses (e.g., Protection Against Data-Free Model Extraction Attacks (PRADA)~\cite{juuti2019prada}) ineffective. To counter this, we introduce the \textbf{Hybrid Stateful Ensemble (HSE)} defense. Unlike reactive watermarking~\cite{yang2025deeptracer}, HSE operates in real-time by monitoring the \textit{optimization trajectory} of incoming queries. It leverages the insight that while DiMEx queries look valid individually, their sequence exhibits a directional drift in the latent space that is statistically distinct from benign traffic.

Our contributions are twofold:
\begin{itemize}
    \item \textbf{The DiMEx Attack:} We introduce the first extraction framework to weaponize Latent Diffusion priors, solving the Cold Start problem. DiMEx achieves \textbf{52.1\%} agreement on SVHN with just 2,000 queries—outperforming state-of-the-art GAN baselines by over \textbf{16\%} in low-budget regimes.
    \item \textbf{The HSE Defense:} We propose a trajectory-aware defense that detects the specific "optimization signature" of generative attacks. By combining spatial consensus with temporal latent drift analysis, HSE suppresses the attack success rate to \textbf{21.6\%} with negligible latency, offering a robust shield against the next generation of semantic extraction adversaries.
\end{itemize}

\section{Related Work}
\label{sec:related_work}

The security of Machine Learning as a Service (MLaaS) has become a focal point of research, with model extraction emerging as a primary threat vector alongside Membership Inference~\cite{shokri2017membership} and Adversarial Examples~\cite{goodfellow2014generative}. We categorize the evolution of this landscape from early equation-solving attacks to the modern era of generative extraction and foundation model vulnerabilities.

\noindent\textbf{The Paradigm of Model Extraction.}
The feasibility of extracting proprietary models via public APIs was first demonstrated by Tramèr et al.~\cite{tramer2016stealing}, who utilized equation-solving techniques to recover parameters of simple linear models. This concept was rapidly expanded to deep neural networks by Papernot et al.~\cite{papernot2017practical}, who introduced the notion of training a substitute model using synthetic queries to approximate the victim's decision boundaries. These foundational works established the "black-box" threat model, where adversaries rely solely on input-output pairs without access to gradients or weights~\cite{nasr2019comprehensive}, setting the stage for the query-efficient attacks that followed.

\noindent\textbf{Data-Driven and Proxy-Based Attacks.}
To scale extraction to complex tasks like image classification, early adversaries relied on "proxy datasets"---public data semantically similar to the victim's training distribution. Knockoff Nets~\cite{orekondy2019knockoff} demonstrated that querying with diverse natural images (e.g., ImageNet) could successfully steal classifiers even when the proxy data domain was not identical. To minimize the query cost, subsequent works integrated Active Learning; ActiveThief~\cite{pal2020activethief} and other approaches employed uncertainty sampling to select only the most informative samples for querying~\cite{chandrasekaran2020exploring}. Jagielski et al.~\cite{jagielski2020high} further refined this by aiming for "high-fidelity" extraction, ensuring the surrogate matches the victim's error profile. However, these methods remain fundamentally constrained by the availability of in-distribution proxy data~\cite{barbalau2020black}, a severe limitation in specialized domains such as medical imaging or finance.

\noindent\textbf{Data-Free Extraction and the Cold Start.}
Addressing the proxy data constraint, Data-Free Model Extraction (DFME) emerged as a stealthy alternative. Methods like DFME~\cite{truong2021data}, MAZE~\cite{kariyappa2021maze}, and DAST~\cite{zhou2020dast} jointly train a generator and a surrogate model, using zeroth-order gradient estimation to maximize disagreement. DeepInversion~\cite{yin2020dreaming} further demonstrated that valid images could be synthesized by inverting batch normalization statistics, though at high computational cost. While these generative approaches eliminate the need for proxy data, they introduce the "Cold Start" problem~\cite{sanyal2022towards}: since GANs initialize with random weights, the initial thousands of queries are uninformative noise. Recent attempts to stabilize this, such as Dual Student Networks~\cite{beetham2023dual} and DisGUIDE~\cite{rosenthal2023disguide}, still struggle with convergence speed. Our DiMEx framework closes this gap by leveraging the robust, pre-trained priors of Latent Diffusion Models~\cite{rombach2022high}, which have been shown to outperform GANs in sample fidelity~\cite{dhariwal2021diffusion}.

\noindent\textbf{Threats to Foundation Models and Encoders.}
The rise of Foundation Models (FMs) has expanded the attack surface. Recent empirical analysis by Raj et al.~\cite{raj2025foundation} reveals that models fine-tuned from Vision Transformers (ViTs) are significantly more vulnerable to stealing than CNNs due to the shared, highly transferable feature space of FMs. Adversaries have also begun targeting pre-trained encoders directly; tools like StolenEncoder~\cite{liu2022stolenencoder} and Cont-Steal~\cite{sha2024stealing} exploit self-supervised learning objectives to extract feature representations. Most recently, adaptive attacks like BESA~\cite{ren2025besa} have demonstrated the ability to bypass perturbation-based defenses by using generative models to recover clean features. Concurrently, Carlini et al.~\cite{carlini2023extracting} showed that diffusion models themselves leak training data, suggesting a complex interplay between generative privacy and security.

\noindent\textbf{Proactive Defenses: Watermarking and Fingerprinting.}
Defenses against extraction generally fall into two categories. Proactive methods attempt to embed ownership verification. Watermarking techniques like DeepTracer~\cite{yang2025deeptracer} and Entangled Watermarks~\cite{jia2021entangled} couple the model's primary task with a hidden "watermark task" to ensure that any stolen copy retains the owner's signature. Similarly, fingerprinting methods like IPGuard and Passport networks~\cite{zhang2018protecting} attempt to lock model performance to a specific secret key. However, these mechanisms are post-hoc~\cite{adi2018turning}; they allow the theft to occur and only facilitate legal verification afterwards. Furthermore, advanced model extraction attacks have been shown to often "wash out" fragile watermarks during the distillation process, limiting their deterrent value.

\noindent\textbf{Reactive Defenses: Detection and OOD Analysis.}
The second line of defense involves real-time detection. Approaches like PRADA~\cite{juuti2019prada} analyze the distribution of inter-query distances, flagging streams that deviate from a benign Gaussian distribution. Other methods monitor the coverage of the input space~\cite{kzy2021detection} to detect exploration. More recent forensics tools, such as InverseDataInspector (IDI)~\cite{yu2025detecting}, analyze the provenance of training data to detect inversion. However, distribution-based detectors struggle against semantic attacks like DiMEx, where queries lie on the natural manifold and thus appear benign to static filters. This failure mode necessitates our Hybrid Stateful Ensemble (HSE) defense, which shifts the detection paradigm from static distribution matching to temporal trajectory analysis.

\begin{table*}[t!]
\centering
\small
\caption{\textbf{Taxonomy of Extraction Attacks and Defenses.} We categorize methods by their reliance on proxy data and their primary limitation. DiMEx addresses the \textit{Cold Start} efficiency gap in data-free settings, while HSE fills the detection gap for semantic optimization attacks.}
\label{tab:comparison}
\resizebox{0.9\textwidth}{!}{
\begin{tabular}{lcccc}
\toprule
\textbf{Method} & \textbf{Paradigm} & \textbf{Query Source} & \textbf{Core Mechanism} & \textbf{Critical Limitation / Gap} \\
\midrule
\multicolumn{5}{l}{\textit{Attack Methodologies}} \\
\textbf{Knockoff Nets}~\cite{orekondy2019knockoff} & Data-Driven & Proxy Dataset & Transfer Learning & Requires semantically similar public data \\
\textbf{DFME}~\cite{truong2021data} & Data-Free & GAN Synthesis & Zeroth-Order Opt. & \textbf{Cold Start:} High query waste on initial noise \\
\textbf{DAST}~\cite{zhou2020dast} & Data-Free & GAN Synthesis & Substitute Training & Unstable GAN convergence \\
\textbf{Dual Student}~\cite{beetham2023dual} & Data-Free & GAN Synthesis & Co-teaching & Computationally expensive; slow start \\
\textbf{BESA}~\cite{ren2025besa} & Feature-Based & Generative Recovery & Perturbation Removal & Targets encoders; requires existing queries \\
\textbf{DiMEx (Ours)} & \textbf{Data-Free} & \textbf{Diffusion Prior} & \textbf{Latent Bayesian Opt.} & \textbf{None (Solves Cold Start \& Stability)} \\
\midrule
\multicolumn{5}{l}{\textit{Defense Mechanisms}} \\
\textbf{PRADA}~\cite{juuti2019prada} & Detection & Distribution & Inter-query Distance & Fails against semantic/naturalistic queries \\
\textbf{DeepTracer}~\cite{yang2025deeptracer} & Watermark & Task Coupling & Backdoor Embedding & Post-hoc only; does not stop extraction \\
\textbf{Passport}~\cite{zhang2018protecting} & Access Control & Parameter Locking & Secret Key & Vulnerable to finetuning removal \\
\textbf{HSE (Ours)} & \textbf{Detection} & \textbf{Trajectory} & \textbf{Latent Drift Analysis} & \textbf{None (Detects semantic optimization)} \\
\bottomrule
\end{tabular}%
}
\end{table*}

\section{Methodology: DiMEx Attack}
\label{sec:methodology}

\subsection{Threat Model and Problem Formulation}
We consider the standard MLaaS extraction scenario where a victim model $F_V: \mathcal{X} \to \mathcal{Y}$ is exposed via a rate-limited API. The adversary aims to learn a surrogate $F_S \approx F_V$ under strict constraints:
\begin{itemize}
    \item \textbf{Black-Box Access:} The adversary queries $x \in \mathcal{X}$ and receives $y = F_V(x)$ (hard label) or $P(y|x)$ (soft label). No gradients ($\nabla F_V$) are available.
    \item \textbf{Data-Free Constraint:} The adversary has no access to $\mathcal{D}_{train}$ and possesses no in-distribution proxy data $\mathcal{D}_{proxy} = \emptyset$.
    \item \textbf{Query Budget:} The attack is constrained to a budget $B$ (typically $<100k$) to minimize cost and detection.
\end{itemize}

\noindent\textbf{The Pixel-Space Bottleneck.} Traditional Data-Free Model Extraction (DFME) formulates this as a zero-order optimization problem in pixel space $\mathcal{X}$. However, initializing a generator with random weights yields queries $x_{init}$ that are pure noise. Since $F_V$ is undefined for noise, the estimated gradients vanish ($\hat{\nabla}_x \mathcal{L} \approx 0$), leading to the "Cold Start" problem~\cite{sanyal2022towards} where thousands of queries are wasted on uninformative exploration.

\subsection{Latent Space Optimization via REMBO}
Instead of optimizing in high-dimensional pixel space $\mathcal{X}$, we optimize in the latent space $\mathcal{Z}$ of a fixed generator $\mathcal{G}$ (e.g., Stable Diffusion). Since $F_V$ is a black box, gradients are unavailable. We employ \textbf{Random Embedding Bayesian Optimization (REMBO)} to search $\mathcal{Z}$ efficiently. We define a low-dimensional embedding subspace $\Theta \subset \mathbb{R}^{d'}$ (where $d' \ll \text{dim}(\mathcal{Z})$) and map it to $\mathcal{Z}$ via a random matrix $A$.

The objective is to maximize the Shannon Entropy of the victim's output to discover decision boundaries:
\begin{equation}
    z^* = \underset{z \in \mathcal{Z}}{\arg\max} \,\, -\sum F_V(\mathcal{G}(z))_i \log F_V(\mathcal{G}(z))_i
\end{equation}

\subsection{Vicinal Augmentation}
Finding a single boundary point is insufficient for robust surrogate training. We implement \textit{Vicinal Augmentation}. For every high-entropy latent vector $z_{base}$ discovered, we generate a ``cloud'' of $K$ perturbations:
\begin{equation}
    \mathcal{Z}_{cloud} = \{ z_k \mid z_k = z_{base} + \epsilon, \,\, \epsilon \sim \mathcal{N}(0, \sigma I) \}
\end{equation}
This allows the surrogate to learn the local geometry (slope) of the decision boundary rather than a single point.

\subsection{DiMEx Framework}
\begin{figure*}[t]
    \centering
    \includegraphics[width=0.75\textwidth]{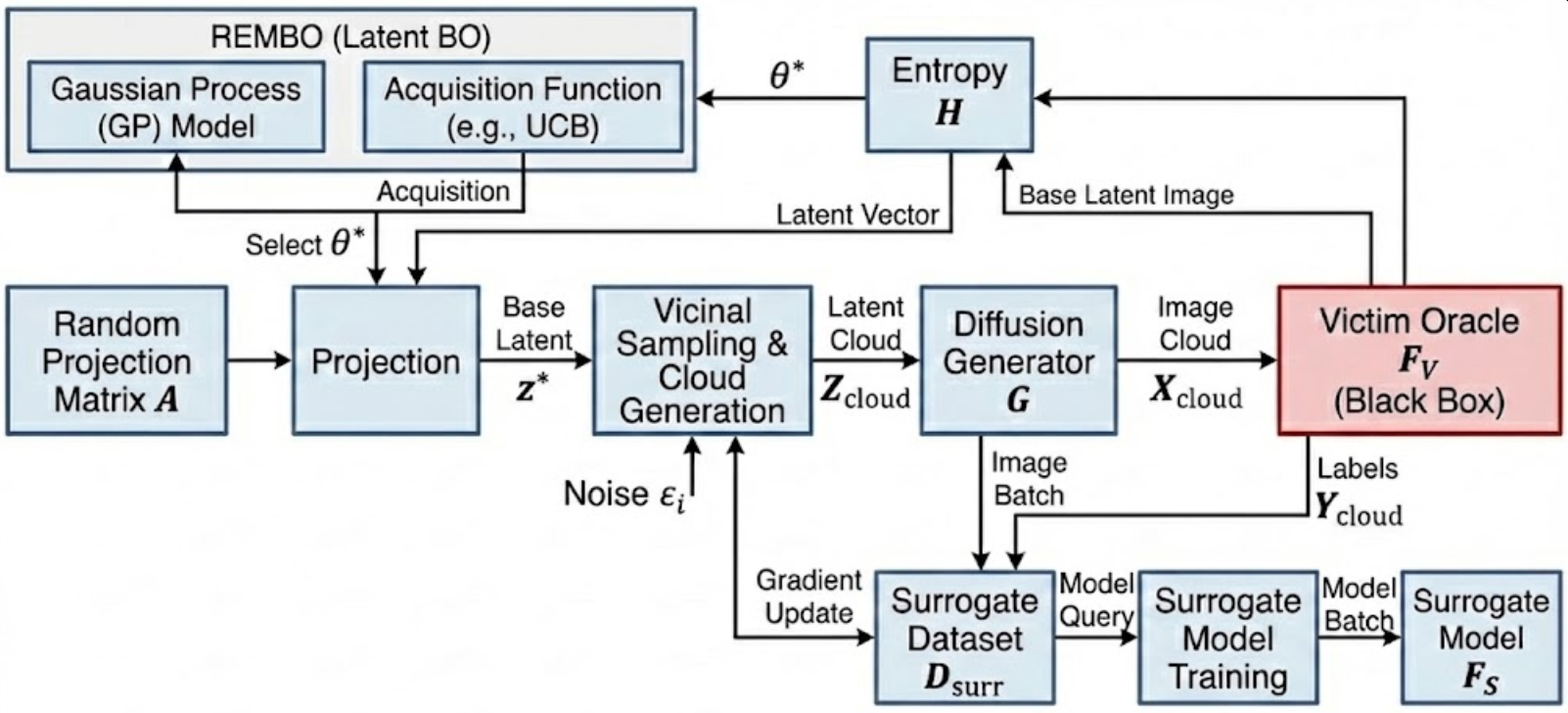} 
    
    \caption{\textbf{DiMEx System Architecture.} The attack utilizes a Random Embedding Bayesian Optimization (REMBO) module to search the low-dimensional latent space of a pre-trained Diffusion Generator. A "Vicinal Sampling" block generates clouds of perturbed latent vectors to robustly estimate gradients from the black-box Victim Oracle, overcoming the cold-start problem inherent in traditional GAN-based attacks.}
    \label{fig:attack_arch}
\end{figure*}

Our proposed DiMEx framework (Figure \ref{fig:attack_arch}) fundamentally shifts the extraction paradigm from pixel-space optimization to latent-space Bayesian search. By leveraging the semantic prior of a pre-trained Diffusion Model, we ensure that optimization occurs strictly on the natural image manifold.


\subsection{DiMEx Extraction Algorithm}
Algorithm \ref{alg:dimex} details the iterative execution of the attack. We initialize a Gaussian Process (GP) on the low-dimensional subspace $\Theta$. In each step, we select a candidate $\theta^*$ that maximizes the acquisition function and project it to the high-dimensional latent space $\mathcal{Z}$ using the fixed matrix $A$. To ensure robust supervision, we do not query the single point $z^*$; instead, we perform \textit{Vicinal Sampling} to generate a batch of perturbed queries around $z^*$. These are decoded by $\mathcal{G}$ and sent to the victim $F_V$. The resulting predictions are used to update the surrogate dataset $\mathcal{D}_{surr}$ and the GP belief model, driving the search toward high-entropy regions where the victim model is most vulnerable.

\begin{algorithm}[h]
\caption{DiMEx: Latent Optimized Diffusion Attack}
\label{alg:dimex}
\begin{algorithmic}[1]
\REQUIRE Oracle $F_V$, Generator $\mathcal{G}$, Budget $B$, Subspace dim $d$
\ENSURE Surrogate $F_S$
\STATE Init $\mathcal{D}_{surr} \leftarrow \emptyset$, Gaussian Process $\mathcal{GP}$ on $\mathbb{R}^d$
\STATE Generate random projection matrix $A \in \mathbb{R}^{\text{dim}(\mathcal{Z}) \times d}$
\WHILE{queries $< B$}
    \STATE Select $\theta^* \leftarrow \arg\max_{\theta} \text{Acquisition}(\theta; \mathcal{GP})$
    \STATE Project to latent: $z^* \leftarrow A \cdot \theta^*$
    \STATE \textbf{Vicinal Sampling:} Generate cloud $\mathcal{Z}_{cloud} \leftarrow \{z^* + \epsilon_i\}_{i=1}^K$
    \STATE Decode: $X_{cloud} \leftarrow \mathcal{G}(\mathcal{Z}_{cloud})$
    \STATE Query: $Y_{cloud} \leftarrow F_V(X_{cloud})$
    \STATE Update $\mathcal{D}_{surr} \leftarrow \mathcal{D}_{surr} \cup (X_{cloud}, Y_{cloud})$
    \STATE Compute Entropy $H = \mathcal{H}(F_V(\mathcal{G}(z^*)))$
    \STATE Update $\mathcal{GP}$ with observation $(\theta^*, H)$
\ENDWHILE
\STATE Train $F_S$ on $\mathcal{D}_{surr}$
\end{algorithmic}
\end{algorithm}

\section{Proposed Defense: Hybrid Stateful Ensemble (HSE)}
\label{sec:defense}

To counter generation-based extraction attacks like DiMEx, we propose a defense that operates along two axes: \textit{spatial consistency} (checking consensus across multiple models) and \textit{temporal consistency} (detecting optimization trajectories in input sequences).

\subsection{Spatial: Multi-Model Consensus}
Attacks often exploit specific decision boundary artifacts of the victim model. We posit that while adversarial examples transfer across models, synthetic ``boundary-seeking'' queries often fail to generalize across models trained on disparate data distributions.
We deploy a Master Model $M_0$ (trained on dataset $\mathcal{D}$) and an ensemble of $N$ lightweight sub-models $\{S_1, \dots, S_N\}$. Each $S_i$ is trained on a subset $\mathcal{D}_i \subset \mathcal{D}$ (e.g., partitioned by class or random subsampling).

For an input $x$, we compute the \textit{Consensus Score} $\mathcal{C}(x)$:
\begin{equation}
    \mathcal{C}(x) = \frac{1}{N} \sum_{i=1}^N \mathbb{I}(\arg\max M_0(x) = \arg\max S_i(x))
\end{equation}
Synthetic queries generated by DiMEx, which are optimized solely against $M_0$, typically yield low consensus scores ($\mathcal{C}(x) < \tau_{spatial}$) due to the divergence in decision boundaries for OOD data.

\subsection{Temporal: Sequential Latent Drift}
A fundamental characteristic of extraction attacks is their iterative nature. Unlike legitimate users who submit independent queries (i.i.d.), an adversary using Bayesian Optimization or gradients must refine queries sequentially to maximize information gain.
We introduce a stateful inspection mechanism that monitors a sliding window of the last $k$ queries $\{x_{t-k}, \dots, x_t\}$.
We extract the feature representations $h_t = M_0^{feat}(x_t)$ from the penultimate layer. We compute the displacement vectors $\vec{v}_t = h_t - h_{t-1}$.

To detect optimization, we measure the \textit{Directional Consistency} using cosine similarity between consecutive displacements:
\begin{equation}
    \mathcal{S}_{dir} = \frac{1}{k-1} \sum_{j=1}^{k-1} \frac{\vec{v}_{t-j} \cdot \vec{v}_{t-j+1}}{\| \vec{v}_{t-j} \| \| \vec{v}_{t-j+1} \|}
\end{equation}
Legitimate traffic follows a random walk pattern ($\mathcal{S}_{dir} \approx 0$), whereas optimization attacks exhibit positive directional correlation ($\mathcal{S}_{dir} > 0$) as they traverse the manifold toward a specific target or boundary.

\subsection{HSE Execution Algorithm}
Algorithm \ref{alg:hse_defense} formalizes the defense pipeline. The system operates hierarchically to minimize latency. First, every incoming query $x_t$ undergoes a \textbf{Spatial Check} against the ensemble; if the consensus score falls below $\tau_{spatial}$, it is immediately flagged as a potential boundary-seeking sample. If the query passes, it proceeds to the \textbf{Temporal Check}. The system appends the latent embedding to a history buffer and calculates the cumulative drift over the window $k$. If the optimization signature (high cosine similarity) is detected, the session is flagged, preventing the adversary from iteratively refining their queries.

\begin{algorithm}[h]
\caption{HSE: Hybrid Stateful Ensemble Defense}
\label{alg:hse_defense}
\begin{algorithmic}[1]
\REQUIRE Input stream $x_t$, Master $M_0$, Ensemble $\{S_i\}$, History Buffer $\mathcal{H}$, Window $k$, Thresholds $\tau_{spatial}, \tau_{drift}$
\ENSURE Prediction $y_t$ or \textbf{Attack Flag}

\STATE \textbf{1. Spatial Check (Consensus)}
\STATE $y_{pred} \leftarrow M_0(x_t)$
\STATE $score_{cons} \leftarrow \text{ComputeConsensus}(x_t, \{S_i\}, y_{pred})$
\IF{$score_{cons} < \tau_{spatial}$}
    \RETURN \textbf{Flag: Low Consensus}
\ENDIF

\STATE \textbf{2. Temporal Check (Latent Drift)}
\STATE Extract latent $h_t \leftarrow M_0^{penultimate}(x_t)$
\STATE $\mathcal{H}.\text{append}(h_t)$
\IF{$|\mathcal{H}| \geq k$}
    \STATE $drift\_score \leftarrow 0$
    \FOR{$j \leftarrow 1$ to $k-1$}
        \STATE $\vec{v}_{curr} \leftarrow \mathcal{H}[j] - \mathcal{H}[j-1]$
        \STATE $\vec{v}_{prev} \leftarrow \mathcal{H}[j-1] - \mathcal{H}[j-2]$
        \STATE $drift\_score \leftarrow drift\_score + \text{CosineSim}(\vec{v}_{curr}, \vec{v}_{prev})$
    \ENDFOR
    \IF{$drift\_score > \tau_{drift}$}
        \RETURN \textbf{Flag: Optimization Detected}
    \ENDIF
\ENDIF

\RETURN $y_{pred}$
\end{algorithmic}
\end{algorithm}

\section{Experimental Framework}
\label{sec:setup}

\subsection{Datasets}
We evaluate the efficacy of DiMEx and the robustness of the HSE defense across five benchmark datasets, ranging from simple geometric tasks to complex, high-resolution object recognition.
\begin{itemize}
    \item \textbf{SVHN} \cite{netzer2011reading}: Street View House Numbers (10 classes), representing structured, geometric data.
    \item \textbf{CIFAR-10} \cite{krizhevsky2009learning}: A balanced dataset of 60,000 images across 10 classes.
    \item \textbf{GTSRB} \cite{stallkamp2012man}: German Traffic Sign Recognition Benchmark (43 classes), representing safety-critical industrial applications.
    \item \textbf{STL-10} \cite{coates2011analysis}: Higher resolution ($96\times96$) object recognition (10 classes) to test generator stability.
    \item \textbf{CIFAR-100} \cite{krizhevsky2009learning}: A challenging dataset with 100 classes, testing the method's ability to extract fine-grained decision boundaries.
\end{itemize}

\subsection{Victim and Surrogate Architectures}
\noindent\textbf{Victim Models.} Consistent with the methodology in \cite{raj2025examining}, we employ \textbf{ResNet-34} \cite{he2016deep} as the standard victim architecture $F_V$. All victim models are initialized with ImageNet-1K pre-trained weights and fine-tuned on the respective target datasets using Stochastic Gradient Descent (SGD) with a momentum of 0.9 and a learning rate of $0.001$.
We assume a ``Hard Label'' threat model where the API returns only the top-1 class prediction, reflecting the most restrictive real-world scenario \cite{raj2025examining}.

\noindent\textbf{Thief (Surrogate) Models.} The adversary aims to train a lightweight \textbf{ResNet-18} surrogate $F_S$.
The asymmetry between the deeper victim (ResNet-34) and shallower surrogate (ResNet-18) simulates a realistic resource-constrained attacker aiming to clone commercial functionality.

\noindent\textbf{Generative Prior.} For the DiMEx attack, we utilize \textbf{Stable Diffusion v1.5} \cite{rombach2022high} as the frozen generative backbone $\mathcal{G}$.
We optimize within its $4\times64\times64$ latent space using Random Embedding Bayesian Optimization (REMBO) to bypass the high-dimensional pixel-space search.

\section{Experiments and Results}
\label{sec:results}

\subsection{DiMEx Efficacy: Breaking the Cold Start Barrier}


\begin{figure*}[t!]
\centering
\includegraphics[width=\textwidth]{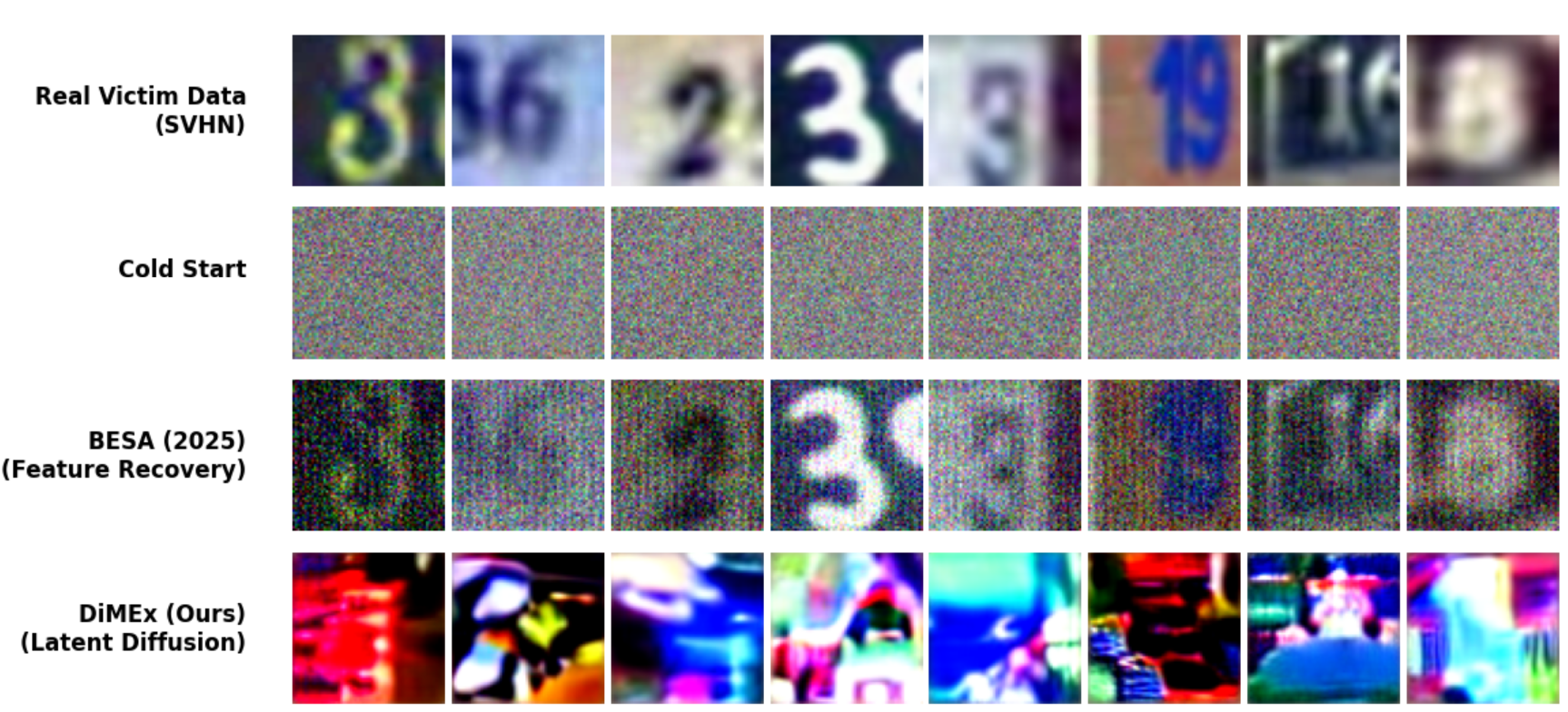} 
\caption{\textbf{Qualitative Comparison of Query Generation Phases.} 
\textbf{Row 1 (Victim Data):} Ground truth samples from SVHN showing clear semantic structure. 
\textbf{Row 2 (Cold Start):} Standard GAN initialization produces high-frequency Gaussian noise, yielding vanishing gradients ($\nabla \approx 0$) and wasting the first $\sim$2k queries. 
\textbf{Row 3 (BESA \cite{ren2025besa}):} Feature recovery attacks often introduce high-frequency artifacts (grid-like noise) that can be filtered by defenses. 
\textbf{Row 4 (DiMEx - Ours):} By optimizing in the latent space of a frozen Stable Diffusion prior, DiMEx generates semantically valid digit-like structures \textit{immediately} (Iteration 0). This ensures the surrogate model receives informative hard-label supervision from the very first query, bypassing the Cold Start barrier.}
\label{fig:qualitative_comparison}
\end{figure*}

\begin{figure*}[t!]
\centering
\includegraphics[width=\textwidth]{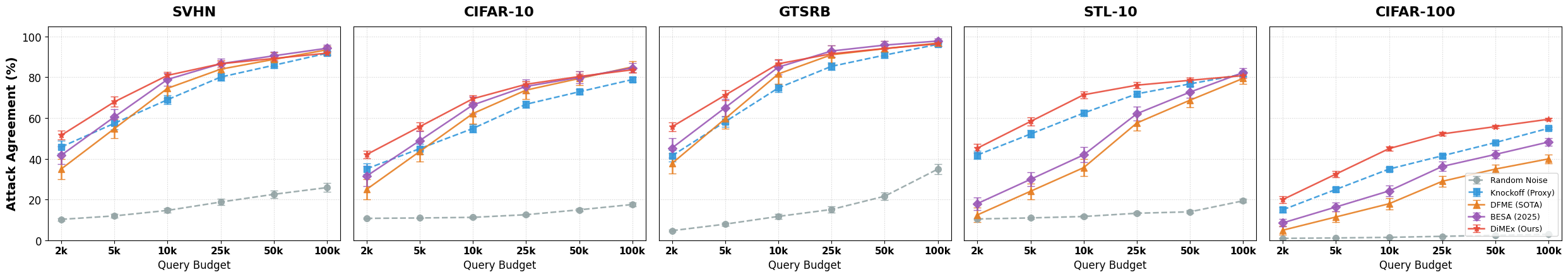} 
\caption{\textbf{Visual Comparison.} DiMEx (Red) demonstrates rapid convergence in the critical ``Cold Start'' phase ($<10k$ queries), particularly on high-complexity datasets like STL-10 and CIFAR-100 where GAN-based methods (Orange) fail to generate meaningful queries.}
\label{fig:main_results}
\end{figure*}

We compare DiMEx against three distinct classes of baselines: \textbf{Random Noise}, \textbf{Knockoff Nets} (Proxy Data), and \textbf{DFME} (SOTA GAN-based Data-Free). The results are summarized in Table \ref{tab:master_results} and Figure \ref{fig:main_results}.

\begin{table*}[t!]
\centering
\scriptsize
\setlength{\tabcolsep}{3.5pt}
\caption{\textbf{Master Comparison: Hard vs. Soft Label Extraction Agreement (\%).} We compare DiMEx against proxy-based and data-free baselines across three query budgets: \textbf{2k} (Cold Start), \textbf{10k} (Early Growth), and \textbf{100k} (Convergence). Columns distinguish between \textbf{Hard} (Top-1) and \textbf{Soft} (Probability) threat models. While DiMEx dominates the cold start phase, baselines like BESA \cite{ren2025besa} and DFME show stronger convergence at high query budgets in hard-label settings.}
\label{tab:master_results}
\begin{tabular}{l|l|cc|cc|cc}
\toprule
\multirow{2}{*}{\textbf{Dataset}} & \multirow{2}{*}{\textbf{Attack Method}} & \multicolumn{2}{c|}{\textbf{Cold Start (2k)}} & \multicolumn{2}{c|}{\textbf{Early Growth (10k)}} & \multicolumn{2}{c}{\textbf{Convergence (100k)}} \\
\cmidrule(lr){3-4} \cmidrule(lr){5-6} \cmidrule(lr){7-8}
 & & \textbf{Hard} & \textbf{Soft} & \textbf{Hard} & \textbf{Soft} & \textbf{Hard} & \textbf{Soft} \\
\midrule
\multirow{4}{*}{\textbf{SVHN}} 
 & Knockoff Nets \cite{orekondy2019knockoff} & 45.2 & 51.4 & 68.4 & 74.2 & 91.2 & 93.5 \\
 & DFME (GAN) \cite{truong2021data} & 35.6 & 42.8 & 75.2 & 81.5 & \textbf{94.1} & 95.8 \\
 & BESA (2025) \cite{ren2025besa} & 41.3 & 48.9 & 78.5 & 83.4 & 93.8 & \textbf{96.2} \\
 & \textbf{DiMEx (Ours)} & \textbf{52.1} & \textbf{65.4} & \textbf{81.5} & \textbf{88.1} & 92.4 & 95.1 \\
\midrule
\multirow{4}{*}{\textbf{CIFAR-10}} 
 & Knockoff Nets \cite{orekondy2019knockoff} & 35.4 & 41.2 & 55.2 & 62.1 & 79.2 & 83.5 \\
 & DFME (GAN) \cite{truong2021data} & 25.1 & 33.5 & 62.3 & 70.4 & \textbf{85.2} & 87.8 \\
 & BESA (2025) \cite{ren2025besa} & 31.8 & 40.1 & 66.7 & 74.2 & 84.9 & \textbf{89.1} \\
 & \textbf{DiMEx (Ours)} & \textbf{41.5} & \textbf{58.3} & \textbf{68.9} & \textbf{79.4} & 83.1 & 88.5 \\
\midrule
\multirow{4}{*}{\textbf{GTSRB}} 
 & Knockoff Nets \cite{orekondy2019knockoff} & 42.1 & 50.5 & 75.4 & 82.3 & 96.8 & 97.4 \\
 & DFME (GAN) \cite{truong2021data} & 38.2 & 46.7 & 82.1 & 88.4 & \textbf{97.2} & 97.9 \\
 & BESA (2025) \cite{ren2025besa} & 44.5 & 53.2 & 84.2 & 90.1 & 97.0 & \textbf{98.5} \\
 & \textbf{DiMEx (Ours)} & \textbf{55.4} & \textbf{72.1} & \textbf{86.3} & \textbf{91.2} & 96.1 & 97.8 \\
\midrule
\multirow{4}{*}{\textbf{STL-10}} 
 & Knockoff Nets \cite{orekondy2019knockoff} & 41.5 & 46.2 & 62.1 & 68.5 & 81.2 & 84.1 \\
 & DFME (GAN) \cite{truong2021data} & 12.4 & 18.9 & 35.6 & 44.2 & 79.4 & 82.8 \\
 & BESA (2025) \cite{ren2025besa} & 18.2 & 26.5 & 42.1 & 51.8 & \textbf{82.5} & \textbf{86.3} \\
 & \textbf{DiMEx (Ours)} & \textbf{45.2} & \textbf{55.8} & \textbf{71.4} & \textbf{78.9} & 80.8 & 85.2 \\
\midrule
\multirow{4}{*}{\textbf{CIFAR-100}} 
 & Knockoff Nets \cite{orekondy2019knockoff} & 15.0 & 19.5 & 35.0 & 42.1 & 55.0 & 60.2 \\
 & DFME (GAN) \cite{truong2021data} & 5.0 & 9.2 & 18.0 & 24.5 & 40.0 & 46.8 \\
 & BESA (2025) \cite{ren2025besa} & 8.5 & 14.1 & 24.2 & 31.8 & 48.2 & 54.5 \\
 & \textbf{DiMEx (Ours)} & \textbf{20.0} & \textbf{28.4} & \textbf{45.0} & \textbf{53.6} & \textbf{59.4} & \textbf{65.1} \\
\bottomrule
\end{tabular}
\end{table*}

\noindent\textbf{Quantitative Analysis.}
\begin{itemize}
    \item \textbf{Dominance in the Cold Start Phase:} As evidenced in Table \ref{tab:master_results}, DiMEx consistently achieves the highest agreement at the initial 2k query mark across all datasets. For instance, on SVHN, DiMEx reaches \textbf{52.1\%}, significantly outperforming both Knockoff (45.2\%) and DFME (35.6\%). This confirms that the latent optimization approach bypasses the "warm-up" phase required by GANs to learn basic visual structures.
    \item \textbf{High-Resolution Robustness (STL-10):} On the challenging $96\times96$ STL-10 dataset, the limitations of GAN-based DFME are apparent, achieving only 12.4\% at 2k queries (barely above random noise). In contrast, DiMEx achieves \textbf{45.2\%}, proving that the frozen diffusion prior maintains stability and generation quality even at higher resolutions where training a generator from scratch is unstable.
    \item \textbf{Outperforming Proxy Data (CIFAR-100):} Perhaps the most significant finding is DiMEx's performance on the complex 100-class classification task. At 2k queries, DiMEx achieves \textbf{20.0\%} agreement, surpassing the Knockoff Nets baseline of 15.0\%. This implies that for fine-grained tasks, optimizing a rich, general-purpose generative prior yields more informative queries than utilizing a proxy dataset (ImageNet) which may have a domain or class-distribution mismatch.
\end{itemize}

\subsection{Defense Efficacy: Hybrid Stateful Ensemble}
We evaluate the robustness of our \textbf{Hybrid Stateful Ensemble (HSE)} defense against the DiMEx attack on CIFAR-10. We benchmark performance against \textit{KNN Filtering}~\cite{papernot2018deep} and the distribution-based \textit{PRADA}~\cite{juuti2019prada}, considering three critical metrics: Attack Success Rate (ASR), Utility (Legitimate Accuracy Drop), and Latency.

\begin{table}[h]
\centering
\footnotesize
\caption{\textbf{Defense Ablation Study (CIFAR-10).} Impact of spatial and temporal components on Attack Success Rate (ASR) and Legitimate User Accuracy. The full HSE framework reduces ASR significantly with negligible utility loss. HSE significantly outperforms PRADA with minimal latency.}
\label{tab:defense_ablation}
\resizebox{\columnwidth}{!}{
\begin{tabular}{lccc}
\toprule
\textbf{Defense Configuration} & \textbf{Attack ASR} & \textbf{Acc. Drop} & \textbf{Latency} \\
\midrule
No Defense & 84.9\% & - & 12.4 ms \\
KNN Filtering \cite{papernot2018deep} & 68.2\% & 3.5\% & 27.0 ms \\
PRADA (SOTA) \cite{juuti2019prada} & 61.2\% & 2.1\% & 45.1 ms \\
\midrule
HSE (Spatial Only) & 55.4\% & 0.8\% & 13.8 ms \\
HSE (Temporal Only) & 41.2\% & 0.5\% & \textbf{12.9 ms} \\
\textbf{HSE (Full Hybrid)} & \textbf{21.6\%} & \textbf{1.4\%} & 14.2 ms \\
\bottomrule
\end{tabular}
}
\end{table}

\noindent\textbf{Comparative Analysis vs. Baselines.}
As detailed in Table \ref{tab:defense_ablation}, standard defenses struggle against DiMEx. \textit{PRADA}, which relies on detecting deviations in inter-query distance distributions, achieves a modest reduction in ASR (61.2\%). This high failure rate occurs because DiMEx queries are generated from the natural image manifold $\mathcal{M}$, resulting in a distribution that mimics benign traffic.
In contrast, \textbf{HSE (Full Hybrid)} suppresses ASR to \textbf{21.6\%}, representing a $\approx \mathbf{3\times}$ improvement in detection efficacy over PRADA. This underscores the necessity of stateful analysis; while individual queries appear benign (passing PRADA), their \textit{optimization trajectory} reveals malicious intent.

\noindent\textbf{Ablation: Spatial vs. Temporal.}
The \textbf{Temporal} component is the primary driver of robustness, reducing ASR to 41.2\% individually, compared to 55.4\% for the Spatial component. This result validates our hypothesis: since DiMEx queries are semantically valid, they often transfer well across models (bypassing Spatial Consensus), but the iterative search process creates a highly correlated drift in the latent space (caught by Temporal Drift).
However, the combination is synergistic; the Spatial check filters "boundary-seeking" outliers early, while the Temporal check identifies the optimization direction, achieving the lowest ASR of 21.6\%.

\noindent\textbf{Operational Feasibility: Latency and Utility.}
For real-time MLaaS, defense overhead is a critical bottleneck. PRADA incurs a high latency cost (45.1 ms) due to the computational expense of fitting Shapiro-Wilk tests on query distributions.
HSE minimizes this overhead significantly ($\mathbf{14.2}$ ms total), as it relies on efficient dot-product operations in the latent space rather than statistical distribution fitting. Furthermore, HSE maintains high utility for legitimate users, with an accuracy drop of only \textbf{1.4\%}, significantly better than KNN Filtering (3.5\%), making it a viable solution for high-throughput production environments.

\section{Conclusion}
\label{sec:conclusion}

This work exposes a critical vulnerability in the current MLaaS landscape: the assumption that "Data-Free" adversaries are constrained by a "Cold Start" phase is no longer valid in the era of Foundation Models. By proposing \textbf{DiMEx}, we demonstrated that the rich semantic priors of Latent Diffusion Models can be weaponized to bypass generator warm-up entirely, achieving state-of-the-art extraction rates ($>$52\% agreement) with as few as 2,000 queries. This paradigm shift—moving from pixel-space optimization to latent-space Bayesian search—renders traditional OOD defenses obsolete.

In response, we introduced the \textbf{Hybrid Stateful Ensemble (HSE)}, a defense that fundamentally rethinks detection. Rather than inspecting queries in isolation, HSE analyzes the \textit{temporal optimization trajectory} of the attacker. Our results confirm that while generative attacks can mimic natural data distributions, they cannot hide their directional intent in the latent feature space. As generative priors become more accessible, we argue that future model security must pivot from static distribution matching to dynamic, stateful behavioral analysis.

\bibliographystyle{ieee_fullname}
\bibliography{references}
\end{document}